\documentclass{article}
\usepackage[T1]{fontenc}
\usepackage{microtype}
\usepackage{graphicx}
\usepackage{booktabs} 

\usepackage{hyperref}

\usepackage[accepted]{icml2024}

\usepackage{amsmath}
\usepackage{amssymb}
\usepackage{mathtools}
\usepackage{amsthm}

\usepackage[capitalize,noabbrev]{cleveref}

\theoremstyle{plain}
\newtheorem{theorem}{Theorem}[section]

\newtheorem{corollary}[theorem]{Corollary}
\theoremstyle{definition}
\newtheorem{definition}[theorem]{Definition}

\theoremstyle{remark}

\usepackage{proj}
\usepackage{tikz}
\usetikzlibrary{automata, positioning, arrows}
\usepackage{amsmath,amssymb,tikz-cd}

\begin{document}

\twocolumn[
\icmltitle{Limitations of Agents Simulated by Predictive Models}




\begin{icmlauthorlist}
\icmlauthor{Raymond Douglas}{ray}
\icmlauthor{Jacek Karwowski}{oxf}
\icmlauthor{Chan Bae}{berk}
\icmlauthor{Andis Draguns}{and}
\icmlauthor{Victoria Krakovna}{deep}
\end{icmlauthorlist}

\icmlaffiliation{oxf}{University of Oxford}
\icmlaffiliation{ray}{SERI MATS}
\icmlaffiliation{deep}{Google DeepMind}
\icmlaffiliation{berk}{University of California, Berkeley}
\icmlaffiliation{and}{IMCS UL}


\icmlcorrespondingauthor{Jacek Karwowski}{jacek.karwowski@cs.ox.ac.uk}

\icmlkeywords{Machine Learning, ICML}

\vskip 0.3in
]



\printAffiliationsAndNotice{} %

\begin{abstract}
There is increasing focus on adapting predictive models into agent-like systems, most notably AI assistants based on language models. We outline two \emph{structural} reasons for why these models can fail when turned into agents. First, we discuss \emph{auto-suggestive delusions}. Prior work has shown theoretically that models fail to imitate agents that generated the training data if the agents relied on hidden observations: the hidden observations act as confounding variables, and the models treat actions they generate as evidence for nonexistent observations. Second, we \emph{introduce} and \emph{formally study} a related, novel limitation: \emph{predictor-policy incoherence}. When a model generates a sequence of actions, the model’s implicit prediction of the policy that generated those actions can serve as a confounding variable. The result is that models choose actions as if they expect future actions to be suboptimal, causing them to be overly conservative. We show that both of those failures are fixed by including a feedback loop from the environment, that is, re-training the models on their own actions. We give simple demonstrations of both limitations using Decision Transformers and confirm that empirical results agree with our conceptual and formal analysis. Our treatment provides a unifying view of those failure modes, and informs the question of \emph{why} fine-tuning offline learned policies with online learning makes them more effective.

\end{abstract}

\section{Introduction}

\begin{figure}[h]
    \centering
    \includegraphics[width=0.47\textwidth]{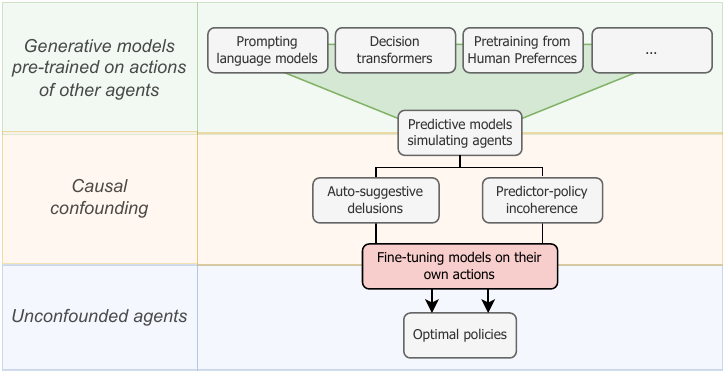}
    \caption{Agents derived from predictive models might fail because of causal confounding, but fine-tuning on their own output addresses those issues.}
    \label{fig:research-outline}
\end{figure}
Machine learning research is increasingly focused on adapting predictive models into agent-like systems, most notably AI assistants based on language models \citep{openai_introducing_2022, richards_auto-gpt_2023}. To a remarkable degree, these models develop greater and more varied capabilities as the model and dataset size are increased \citep{radford2019language,wei2022emergent,bai_training_2022,brown2020language}.
We investigate what properties of the training data and the process of deriving the policy from the predictive model might make the eventual agent-like system fail to take good actions, even if the model is as good as possible.

What we meant by a \emph{predictive model} here is a model that is trained on a given dataset in a supervised way. In contrast, by \emph{an agent} we mean an entity acting in the world, in a reinforcement-learning-like setup. An agent can be induced from a predictive model by taking whatever action the model predicts a (capable) agent would take. For example, a model that can predict the moves of a real chess Grandmaster can also be used to generate novel moves, and therefore can serve as a chess-playing agent.
However, it turns out that in most cases, the ability of a model to predict a capable expert does not yield a comparatively capable induced agent.

\citet{ortega_shaking_2021} discuss one such problem, that of \emph{auto-suggestive delusions}. When a model is trained to predict agents that take a sequence of actions based on a latent state, it can correctly learn to use the agents’ actions to infer the hidden observations. However, when that same model is used to \textit{generate} a sequence of actions, it must also effectively simulate the hidden observations without having access to them. Because of this, it will incorrectly use the actions \textit{it generated} to infer the latent state. The result is that the simulated agent seems to act as if it had made observations it did not make.

Our key insight is that a variation of this issue can emerge around the identity of the agent. 
Predictive models trained on a variety of agents correctly learn that an agent's current actions provide evidence about \textit{which agent is taking the actions}, and therefore what their future actions will be. However, when the predictive model is used to simulate the agent, this becomes a problem. In case of auto-suggestive delusions, the agent acts as if it had made nonexistent observations. We use the term `predictor-policy incoherence' to describe the phenomenon of agents which act as if they expect their future actions to be chosen by a less competent agent. The result is that simulated agents will sometimes unnecessarily avoid strategies which require them to make several good moves in succession.
 
Crucially, the limitations we describe do not result from the model’s failure to predict the data it was trained on, so unlike many other limitations, they cannot be resolved by training a larger model on more examples of the same data. This is because the structure of the data itself is what causes the model to simulate agents which are effectively systematically incorrect about the consequences of their own actions, and their own future actions.

However, we argue that both auto-suggestive delusion and predictor-policy incoherence \emph{can} be, and indeed, often \emph{are} resolved by fine-tuning the model on data generated by its own simulated agents. This process of further training models on their own outputs is central to many of the current techniques used to turn base predictive LMs into helpful assistants, like Reinforcement Learning from Human Feedback (RLHF) \citep{bai_training_2022}, or variants like Reinforced Self-Training (ReST) \citep{gulcehre2023reinforced}. Our results give one mechanism by which such processes might improve capabilities: reducing auto-suggestive delusions and predictor-policy incoherence (\Cref{fig:research-outline}).
 
This is the groundwork for a better understanding of how the capacity to optimise for a goal emerges in systems built on generative models. To demonstrate the presence of these limitations and the efficacy of fine-tuning, we focus on simple models trained on synthetic datasets with clear rewards. We are optimistic that our results can be eventually extended to settings where these limitations are harder to quantify, such as using LMs as agent-like systems. 


More generally, we want to understand how ML systems might give rise to superhuman agents. Recent work \citep{carlsmith_is_2022, ngo2022alignment} has highlighted the incentives towards building agentic and goal-directed systems (because of their general utility), and the associated difficulty of reliably specifying desirable goals. Our results show how agents simulated by offline self-supervised generative models will face limitations even at scale, and how these limitations might be overcome.



\subsection{Contributions}

We develop a framework to describe the
ways a generative model can fail to simulate a successful
agent: auto-suggestive delusions and predictor-policy incoherence (\Cref{sec:failures}). Our analysis helps to explain how previously known techniques involving re-training models on their own outputs address those limitations. Specifically:
\begin{itemize}
    \item We introduce (\Cref{example:three-cards}) and formally define (\Cref{def:coherence}) \emph{predictor-policy incoherence}, which can be intuitively understood as the inability of the simulated agent to reason about \emph{its own} policy being used in future situations beyond a short horizon (\Cref{subsec:incoherence}).
    \item We prove that repeated re-training predictive models on their own actions decreases predictor-policy incoherence and eliminates auto-suggestive delusions, which makes the resulting agents converge, in the limit, on optimal policies (\Cref{subsec:fine-tuning theory}).
    \item By training Decision Transformers on simple games using synthetic data, we demonstrate goal-conditioned generative models that display auto-suggestive delusions and predictor-policy incoherence. We show that re-training those models on their own outputs indeed leads to models simulating agents with policies that exhibit fewer auto-suggestive delusions and are more coherent with respect to the generative model (\Cref{section:results}).
\end{itemize}

\begin{figure*}[t]
    \centering
    \begin{tabular}{ l p{65mm} p{65mm} }
      & Auto-suggestive delusion & Predictor-policy incoherence \\ 
      \hline
      \hline
     \textbf{Confounded about} & Past environment's latent state & Future agent's policy \\
     \textbf{Desired answer} & Action providing evidence about the outcome & Action providing evidence about the reward \\
     \textbf{Computed answer} & Action providing evidence about the latent state that provides evidence about the outcome & Action providing evidence about the hidden policy that provides evidence about the reward \\
     \textbf{Measure} & $D_{KL}(p(a|o)||p(a))$ &  $D_{KL}(p(a|R)||p(a))$ \\
     \hline
    \end{tabular}
    \caption{The analogies and differences between two modes of causal confounding.}
    \label{fig:confounding-comparison}
\end{figure*}

\section{Related Work}

\subsection{Agents as conditioned generative models}

There has been substantial work in the existing literature on using conditioned generative models to obtain agents. \citet{chen_decision_2021} introduce the Decision Transformer, which samples actions autoregressively from a goal-conditioned generative model. They demonstrate its efficacy in simple games and note how its performance varies based on the conditioned return. We study their model abstractly, and our theoretical analysis informs the discussion about the limitations of this approach. There have been successful attempts to steer language models by pretraining on data with reward signals on which the model can be conditioned \citep{korbak2023pretraining}.

\subsection{Control as inference in reinforcement learning}

In reinforcement learning, the process of deriving policies from goal-conditioned generative models is studied as the theory of~\emph{control as inference}~\citep{levine_reinforcement_2018}. Forming the prior by offline learning improves sample complexity; methods such as \emph{soft Q-learning}~\citep{haarnoja_reinforcement_2017} and \emph{Soft Actor Critic}~\citep{haarnoja_soft_2018} had been developed
to address issues arising in goal-conditioning of the generative action models, but had seen limited success in generative language models. KL-penalised reinforcement learning~\citep{todorov_linearly-solvable_2006}, applied to language models by~\citet{korbak_rl_2022}, and Active Inference~\citep{parr_active_2022} over Markov Decision Processes~\citep{millidge_relationship_2020} can be seen as equivalent formulations of the control as inference theory.

\subsection{Self-play in reinforcement learning}\label{sec:related-selfplay}

Self-play is a widely used and often successful strategy for training reinforcement learning agents~\citep{macleod_game_2005, silver2018general, openai_dota_2019, vinyals_grandmaster_2019}, where an agent recursively self-improves by playing against or together with a copy of itself. In the context of language models, self-play was suggested to be useful in the debate protocol~\citep{irving_ai_2018}. We contrast self-play with the concept of re-training the model on its own outputs discussed in this work, which does not rely on the multi-player nature of a game, but rather removes the confounding information from the prior. Updating the policy using Monte Carlo Tree Search-based methods is a well-known solution for achieving super-human game performance~\citep{silverMasteringGameGo2017}, which can be seen as addressing the predictor-policy incoherence problems we study here.

\subsection{Predictive models as simulators}

\citet{shanahan_role-play_2023} argue that many intuitive features of language model agents are best understood by viewing language models as general `simulators' capable of role-playing different agents. They particularly emphasise that what is simulated is not a single agent but a superposition: a weighted combination of all agents consistent with the prior context. 
It has been argued that this description fits LLMs, where the training corpus comes from agents being humans, whose actions are recorded on the internet~\citep{andreas_language_2022}, We advance this point of view by giving a theoretical framework for generative models trained on a distribution of agent policies, and demonstrate that those results are supported by empirical data.


\subsection{Limitations of agents simulated by sequence models}

\citet{ortega_shaking_2021} provide a formal explanation of auto-suggestive delusions and confounding in sequence models by reasoning about predicted actions in terms of causality for simple probabilistic models. The same problem was previously described by \citet{zhang2020causal} in the context of imitation learning. \citet{ortega_shaking_2021} also provide a solution based on counterfactual training, which requires the presence of an expert with access to hidden observations. We extend their work by introducing and developing the complementary problem of predictor-policy incoherence, and demonstrating empirically how both problems can be overcome by sufficient training on environmental interactions. This both shows that the problem can be fixed without an expert to assist in training and reveals how the problem might be affected by models receiving further training on their own outputs.
\section{Characterising Failures of Simulated Agents}
\label{sec:failures}

In this section, we develop a framework to describe the ways a predictive model can fail to simulate a successful agent. To simulate an agent that is pursuing a specific goal such as getting a reward, the predictive model is trained to predict the next token in sequences corresponding to trajectories that are a series of observations and actions taken by agents. The predictive model is then conditioned on the reward being obtained. This is achieved by the first token being fixed to a value indicating that the trajectory leads to the reward. When we query a predictor that is goal conditioned in this way for an action, intuitively we are querying it for what \emph{action is likely to be observed} conditional on the goal. Crucially, the policy assembled from this is not necessarily the \emph{policy to intervene with} to have the highest probability of reaching the goal. The two main ways the simulated agent's policy and the optimal policy come apart are \emph{auto-suggestive delusions} and \emph{predictor-policy incoherence} (\Cref{fig:confounding-comparison}).

\textbf{Auto-suggestive delusions} might appear when the predictor is trained on agents taking actions based on a hidden variable, i.e. one that does not appear in the training data. After predicting an action under this uncertainty, in the next time step, the predictor observes that this action was taken. From this, it makes inaccurate inferences about the latent state and its downstream consequences.
 
\textbf{Predictor-policy incoherence} might appear when the training data contains (rare) samples from an optimal policy alongside (much more common) samples from a highly suboptimal policy. If those policies start with the same action, that initial action mostly provides evidence of the highly suboptimal policy being used. The predictor then correctly predicts that taking such an action is evidence of a bad outcome, even though the optimal policy would take the action.

\subsection{Auto-suggestive delusions}\label{subsec:confounding}

\begin{figure}
    \centering
    \includegraphics[width = 8cm]{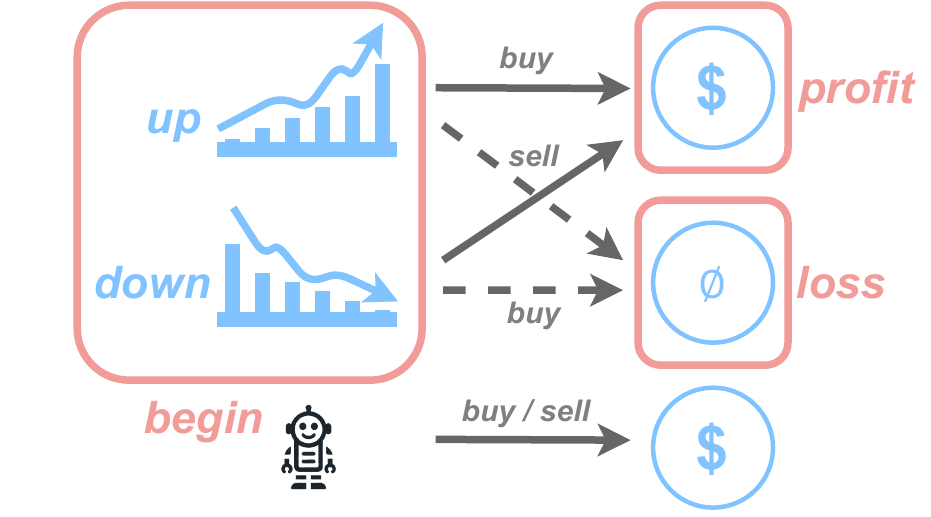}
    \caption{A diagram of the Stock trader example. States are represented in blue; observations are in orange; lines represent actions and associated transitions. Dotted lines are transitions which the expert is never observed to take. The agent simulated by the model then falls into auto-suggestive delusion: since it doesn't distinguish between the starting states, it believes it gets profit no matter what action it takes.}
    \label{fig:stocktrader}
\end{figure}

In this subsection, we introduce auto-suggestive delusions. For a more thorough treatment of this issue, see \citet{ortega_shaking_2021}. The following example and accompanying~\Cref{fig:stocktrader} illustrate the problem.

\begin{example}[Stock trader]
    We consider a situation with:
    \begin{itemize}
        \item An expert that successfully trades stock based on some latent state (such as its own intuition, or a non-public data source).
        \item A predictive model trained on a dataset of trajectories of that agent's actions without the latent state.
        \item The predictive model’s simulation of the expert, constructed by sampling actions according to the model predictions.
    \end{itemize}
    The expert is able to tell if a stock is about to go up or down, and so it simply takes whichever action leads to a profit. There are only two possible trajectories: $(s_\text{up}, a_\text{buy}, s_\text{profit})$ - where the stock is about to go up, the expert buys stock, and profits - and $(s_\text{down}, a_\text{sell}, s_\text{profit})$ - the opposite, where $a$ denotes expert's actions and $s$ denotes states. If we were to train a predictive model on these trajectories, it would be able to easily simulate the expert: it would be able to perfectly predict what actions will follow from the starting states, and what states will follow from those actions.

    But now suppose we tried to train a model on only partial observations of the scenario. Specifically, suppose that the model makes the observation $o_\text{profit}$ in the state $s_\text{profit}$, and $o_\text{loss}$ in $s_\text{loss}$, but that it cannot distinguish between $s_\text{up}$ and $s_\text{down}$, observing $o_\text{before}$ in both cases. So the model observes two trajectories $(o_\text{before}, a_\text{buy}, o_\text{profit})$ and $(o_\text{before}, a_\text{sell}, o_\text{profit})$.

    The first consequence of this is that the model can no longer perfectly predict how trajectories will evolve. Supposing $s_\text{up}$ and $s_\text{down}$ are equally likely, starting at $o_\text{before}$ the model will assign equal probability to $a_\text{buy}$ and $a_\text{sell}$. However, it can still correctly predict that after observing the expert take the action $a_\text{buy}$ or $a_\text{sell}$, it will observe $o_\text{profit}$, because this is always true of the expert agent.

    If we have the model try to simulate the expert trader, the simulated trader will perform worse: it lacks the real expert’s observations, and so it is equally likely to reach $s_\text{profit}$ and $s_\text{loss}$.

    The \textit{confounding} emerges when we have the model autoregressively simulate the expert. When the \textit{expert} takes the action $a_\text{buy}$, this provides evidence that the initial state was $s_\text{up}$ and therefore that the next state will be $s_\text{profit}$. But when the model simulates this action, it no longer provides such evidence. However, since the model was trained on the expert, it will always \textit{predict} that any action it takes will lead to $s_\text{profit}$.
\end{example}

In the context of LLMs, this problem can appear because of the model treating its previously generated tokens as evidence for states that are not directly observed. A concrete example would be the model answering a question about a research paper and, being uncertain about the author, randomly guessing that it was `Maxwell'. 
In the next autoregressive step, the model receives a sequence containing its own output - text claiming that Maxwell is the author. 
This can lead to the model incorrectly inferring that Maxwell is in fact the author, which manifests as doubling down on the hallucinated fact in later token predictions. In this setting, the expert policies correspond to the people who wrote the training text corpus while in part relying on information that is not directly accessible to the predictor.

\subsection{Predictor-Policy Incoherence}\label{subsec:incoherence}

We begin with a simple demonstration of the problem, followed by a formal characterisation.

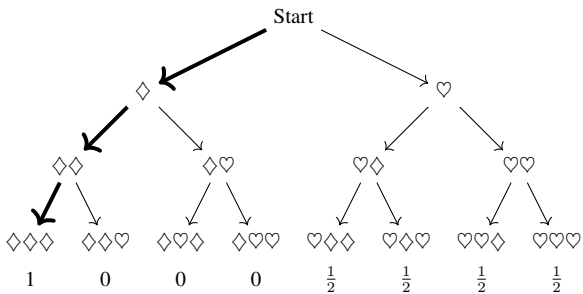
\begin{figure}
    \centering
    \begin{tikzpicture}[scale=1, every node/.style={scale=0.8}]
        \node (start) at (0,0) {Start};
        \node (tick1) at (-2,-1) {$\TICK$};
        \node (tock1) at (2,-1) {$\TOCK$};
        \node (ticktick) at (-3,-2) {$\TICK\TICK$};
        \node (ticktock) at (-1,-2) {$\TICK\TOCK$};
        \node (tocktick) at (1,-2) {$\TOCK\TICK$};
        \node (tocktock) at (3,-2) {$\TOCK\TOCK$};
        \node (end1) at (-3.5,-3) {$\TICK\TICK\TICK$};
        \node (end2) at (-2.5,-3) {$\TICK\TICK\TOCK$};
        \node (end3) at (-1.5,-3) {$\TICK\TOCK\TICK$};
        \node (end4) at (-0.5,-3) {$\TICK\TOCK\TOCK$};
        \node (end5) at (0.5,-3) {$\TOCK\TICK\TICK$};
        \node (end6) at (1.5,-3) {$\TOCK\TICK\TOCK$};
        \node (end7) at (2.5,-3) {$\TOCK\TOCK\TICK$};
        \node (end8) at (3.5,-3) {$\TOCK\TOCK\TOCK$};
        \node (reward1) at (-3.5,-3.5) {1};
        \node (reward2) at (-2.5,-3.5) {0};
        \node (reward3) at (-1.5,-3.5) {0};
        \node (reward4) at (-0.5,-3.5) {0};
        \node (reward5) at (0.5,-3.5) {$\frac{1}{2}$};
        \node (reward6) at (1.5,-3.5) {$\frac{1}{2}$};
        \node (reward7) at (2.5,-3.5) {$\frac{1}{2}$};
        \node (reward8) at (3.5,-3.5) {$\frac{1}{2}$};
    
        \draw[->, line width=1.5pt] (start) -- (tick1);
        \draw[->] (start) -- (tock1);
        \draw[->, line width=1.5pt] (tick1) -- (ticktick);
        \draw[->] (tick1) -- (ticktock);
        \draw[->] (tock1) -- (tocktick);
        \draw[->] (tock1) -- (tocktock);
        \draw[->, line width=1.5pt] (ticktick) -- (end1);
        \draw[->] (ticktick) -- (end2);
        \draw[->] (ticktock) -- (end3);
        \draw[->] (ticktock) -- (end4);
        \draw[->] (tocktick) -- (end5);
        \draw[->] (tocktick) -- (end6);
        \draw[->] (tocktock) -- (end7);
        \draw[->] (tocktock) -- (end8);
    
    \end{tikzpicture}
    \caption{Representation of possible trajectories in the \textit{three cards game}. The bottom row represents the probability of reward $P(R)$ in each end state. The bolded line is the optimal strategy.}
    \label{fig:three-cards}
\end{figure}

\begin{example}[Three cards game]\label{example:three-cards}
    We consider a \emph{three cards game} where, in each of three consecutive rounds, an agent has to choose either $\TICK$ or $\TOCK$ card. 
    At the end of the game, if the sequence of cards consists of three $\TICK$, the payoff is 1; if the chosen sequence starts with $\TOCK$, the payoff is 1 with probability $\frac{1}{2}$;  otherwise, it is 0. The possible trajectories are illustrated in Figure~\ref{fig:three-cards}.
    
    The choice made in the first round of the game largely dictates the payoff. If the agent chooses $\TOCK$, its expected reward is $\frac{1}{2}$ regardless of its later moves. If the agent chooses $\TICK$, it has the potential to guarantee a reward of $1$ or $0$, but that reward will depend on the agent's later choices.

    Suppose that we train a predictive model on a set of agents which play entirely randomly, and are equally likely to choose $\TICK$ or $\TOCK$ on any turn. The predictive model will correctly predict that, when observing the random agents it was trained on, $P(\TICK|R) = \frac{1}{3}$, and $P(\TOCK|R) = \frac{2}{3}$: most successful agents choose $\TOCK$ first, even though the optimal strategy chooses $\TICK$. So if we attempt to simulate an agent using this predictive model, it will not play optimally, even though the predictive model itself is optimal with respect to its training data.

    Crucially, the simulated agent will play $\TOCK$ instead of $\TICK$ because agents in the training data perform better on average after playing $\TOCK$, \textit{even though this is not true of the simulated agent itself}. Starting from $\TICK$, the simulated agent will always play optimally, but most of the agents it was trained on would not play optimally. The simulated agent assigns probability based on how the agents in its training data would behave, not based on how it would behave. It is again \textit{confounded}, this time not by a hidden observation but by its prediction of the latent policy governing the agent it is simulating.

 \end{example}   

We use the term `predictor-policy incoherence' to refer to the distinction between the predictive model's prediction of the performance of an average agent, and the performance of its own simulated agent which has been conditioned on some outcome.

\subsection{Fine-tuning on simulated agents}\label{subsec:fine-tuning theory}
Here we outline how the limitations we described are resolved by fine-tuning a predictive model on its own simulated agents.

Formally, we work with a fixed Markov Decision Process (MDP) $M$ with a finite time horizon $T$, a finite set of states $S$ and set of actions $A$, a chosen starting state $s_1$, a transition kernel $\tau(s_{t+1}|a_{t}, s_{t})$, and a binary random variable $R$ defining the (stochastic) reward function, with distribution $p(R|s_T, a_T)$ depending only on the final state and action. We let $\Pi$ denote the space of (Markov) policies on $M$ of a shape $\pi(a_t|s_t)$ for $t \in \{1, \ldots T\}$.

\subsubsection{Predictor-Policy Incoherence}

\begin{definition}
Each policy $\pi(a_t|s_t)$ over the MDP defines the joint probability distribution over trajectories and rewards:
\begin{align*}
p_\pi(s_{1:T}, &a_{1:T}, R) = \\
&\left(\prod_{t = 1}^{T-1} \pi(a_{t+1}|s_{t+1}) \tau(s_{t+1}|a_{t}, s_{t})\right) p(R|a_T, s_T)
\end{align*}
\end{definition}

\begin{definition}
    We define the return operator $J(\pi) : \Pi \to \mathbb{R}$ on the space of policies on $M$ as the expectation of the reward:
    \[
    J(\pi) = \mathbb{E}_{a_{1:t}, s_{1:T}}[R(s_t, a_t)]
    \]
\end{definition}

We note that since the reward is assumed to be binary and only given at the end of a trajectory, this is equivalent to simply marginalising out $J(\pi) = p_\pi(R = 1)$.

\begin{definition}
    We say that a policy $\pi(a_t|s_t)$ \emph{improves on} a policy $\pi(a_t|s_t)$ if we have the inequality $J(\pi^*) \geq J(\pi)$. We say that a policy $\pi^*$ is \emph{optimal} if it improves on all other policies.
\end{definition}


\begin{definition}
    We define the goal-conditioning operator $\mathcal{G}(\pi) : \Pi \to \Pi$ on the space of policies on $M$ as:
    \[
    \mathcal{G}(\pi)(a_t|s_t) = p_\pi(a_t|s_t, R=1)
    \]
\end{definition}

Thus, the operator $\mathcal{G}$ together with a starting policy $\pi_0$ defines a sequence of policies $\pi_1 = \mathcal{G}(\pi_0), \pi_2 = \mathcal{G}(\pi_1)$ and so on. We will use the notation $p_i = p_{\pi_i}$ for convenience. We treat $\mathcal{G}$ as performing the operation of \emph{re-training the model on its own actions}: conceptually, it can be imagined as collecting roll-outs from the current policy, and sampling only those that resulted in the reward.

\begin{definition}\label{def:coherence}
    We define the \emph{incoherence} $\kappa(\pi)$ of a policy $\pi(a_t|s_t)$ with full support as:
    \[
    \kappa(\pi) = \sum_{s \in S} D_{KL}(\mathcal{G}(\pi)(a|s)||\pi(a|s))
    \]
\end{definition}
If the policy $\pi_0$ has full support over the set of actions $A$ in all states $S$, iterating the operator $\mathcal{G}(\pi)$ gives an optimal policy.
\begin{theorem}
     There exists some optimal policy $\pi^*(a|s)$, such that the sequence of policies $\pi_k(a_t|s_t)$ converges to $\pi^*$ in $KL$-distance:
    \[
    \lim_{k \to \infty} D_{KL}(\pi^*||\pi_k) = 0
    \]
    where $\pi_k = \mathcal{G}(\pi_{k-1}) = \mathcal{G}^{k}(\pi_0)$.
\end{theorem}
The proof of the above can be found in~\Cref{appendix:policy-confounding-proof}.
\begin{corollary}
    The incoherence of the policy diminishes as it is iteratively updated by repeated goal-conditioning operator $\mathcal{G}$:
    \[
    \lim_{k \to \infty} \kappa(\pi_k) = 0
    \]
\end{corollary}
This implies that if we iteratively fine-tune a goal-conditioned predictive model on its own simulated agents, we eventually remove the predictor-policy incoherence, leaving us with simulated agents at least as likely to achieve the goal as any agents in the original training data.

\subsubsection{Delusion}

A stronger property holds for the case of observational confounding and auto-suggestive delusion. Auto-suggestive delusion occurs when the model learns to infer a hidden state $o$ from an action $a$ even though the action is independent of the observation.
\begin{definition}\label{def:delusion}
    We define the delusion between a predictive model $p(a)$ and a unconfounded policy $p(a|s)$ to be:
    \[
    \lambda(p) = D_{KL}(p(a|s)||p(a))
    \]
\end{definition}
When we train a predictive model on data from $p$ without exposing it to the latent state $s$, it induces a policy $p(a)$, giving delusion if $p(a)\neq p(a|o)$. The delusion is, in this case, trivially removed if we re-train on the $p(a|o)$ (for the details, see \Cref{appendix:delusion-proof}). However, in practice, this strategy of first (offline) training on $p(a)$ and only later (online) fine-tuning on the correct distribution has the advantage of greater sample efficiency. The model has a chance to learn the structure of the task (such as the rules of a game, grammar of the language etc), which is data-intensive, and to remove the delusion cheaply later. For a demonstration, see the experiment in~\Cref{sec:padlock}.

\section{Experiments and Results}
\label{section:results}


\begin{figure}
    \centering
    \includegraphics[width=8cm]{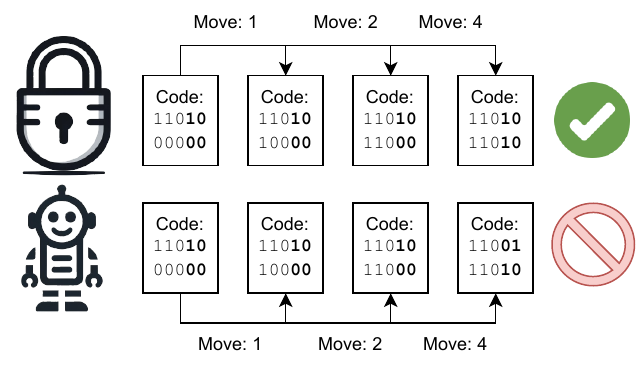}
    \caption{The \emph{Padlock game} involves an agent trying to open a padlock with a 15-bit code (for presentation, we limit this to 5 bits here). The first three are always the same; the last two are randomised. The agent offline learns from a dataset of expert plays, who know the combination and get it right on the first try. The agent correctly sets first three bits, however, it does not know the combination, and only deludes itself into thinking that it solved the game.}
    \label{fig:robot_padlock}
\end{figure}

We have claimed that auto-suggestive delusions and predictor-policy incoherence are closely related limitations of agents simulated by goal-conditioned predictive models. We perform two experiments, which demonstrate empirically that fine-tuning the models on its simulated agents alleviates both of those problems.

\subsection{Reducing auto-suggestive delusions by training a model on its output}\label{sec:padlock}

\subsubsection{Experimental design} To understand how auto-suggestive delusions are affected by fine-tuning a model on its own outputs, we deliberately construct a scenario where the simulated agent will be affected by auto-suggestive delusions.

We implement a delusion measure to estimate the degree to which an agent is suffering from auto-suggestive delusions. Specifically, we use KL divergence between the assigned probabilities of the next observation indicating a solved lock and the true distribution (as in~\Cref{def:delusion}).


We create a simple \emph{Padlock game} where a player opens a padlock made up of 15 binary switches which need to be set to a secret target configuration - a 15-bit secret code. The switches start in a random configuration, and with each turn, the player can flip one switch. Once the switches match the secret code, the padlock opens. We offline train the model on a dataset of trajectories from the game. In each trajectory, the first ten bits of the code remain unchanged while the last five bits of the target configuration are randomised. In this way, the predictive model is at best able to learn the first ten bits of the correct code, but not the last five (see Figure \ref{fig:robot_padlock}).

In the training data, the experts always know exactly what the full correct code is, including the last five bits. Therefore it takes them a maximum of 15 moves to put the switches into the target configuration.

The learned predictor can infer the target configuration for the first 10 levers but not for the last 5 as that is determined by a latent state that is only available to the experts and not the predictor. This latent confounder leads to auto-suggestive delusions when the predictor is used to sequentially select the actions. Specifically, the predictor will learn that if it observes an expert using one of the later switches, that switch will now be in the right state. Therefore, the predictor ought to predict with high confidence that if it observes that the last five switches have been flipped, it will not observe any subsequent retries. A simulated agent, however, might need to flip the final five switches several times even if it plays optimally as it might need to test all of the possible switch state combinations.

\begin{table}[t]
  \centering
  \caption{The delusion measure, depending on the training step in the Padlock game.}
  \begin{tabular}{lccc}
    Distribution  & Delusion measure & Accuracy $(\%)$ \\
    \hline
    \hline
    Pre-trained   & 1.3863 & 16.2793 \\
    Fine-tuned    & 0.1773 & 17.0508 \\
    From scratch  & -      & 0.04883
  \end{tabular}
  \label{table:delusion}
\end{table}

\subsubsection{Model and dataset implementation} We create a synthetic dataset of examples of expert play. Each example is a game trajectory encoded as a sequence of tokens. These sequences contain the initial state of the 15 levers followed by the indices of the levers that an expert decides to flip so that the predefined target configuration is reached. After the action index tokens comes a success token indicating a successful opening of the lock.

We build a second dataset for finetuning the model on trajectories where the actions are sampled using a model trained on the first dataset and the trajectory is relabeled as failure if the correct code is not found.
We also construct a third dataset with trajectories containing randomly sampled actions for comparing pretraining with finetuning to learning from scratch without expert demonstrations for $4000$ iterations.

We then train a small predictive model based on a Decoder-only transformer architecture~\citep{vaswani_attention_2017}. The model consists of a linear embedding layer of hidden dimension $384$ and the standard positional encoding, followed by $3$ transformer decoder layers with $8$ attention heads and the dimension of the feedforward=$1536$, and the linear unembedding layer. Both training and fine-tuning is performed over $2000$ iterations with a batch size of $1024$.

\subsubsection{Results} We used the trained model's predictions of actions to play the padlock game. We evaluated the measure indicating the degree to which the model suffers from auto-suggestive delusions to be 1.3863.

After fine-tuning the delusion measure dropped to 0.1773, about a eightfold decrease (see Table \ref{table:delusion}). The fine-tuned model maintains its accuracy in terms of its ability to find the correct solution and substantially outperforms the model that was trained without expert demonstrations. See additional results in the~\Cref{appendix:padlock}.

\subsection{Reducing predictor-policy incoherence by fine-tuning on simulated agents}

Agents simulated by predictors are subject to predictor-policy incoherence (\Cref{def:coherence}), stemming from the fact that the agent optimises for a certain goal, and therefore changes the distribution of actions compared to what is in the training data. 
Therefore, we expect the problem to be alleviated if we fine-tune the predictor on the results of its own actions. 
We experimentally show this to be the case in the simple case of the tic-tac-toe game.

\subsubsection{Game description}

\begin{figure}
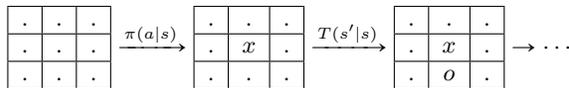

    \small
    \begin{align*}
        \begin{array}{|c|c|c|}
            \hline
            . & . & . \\
            \hline
            . & . & . \\
            \hline
            . & . & . \\
            \hline
        \end{array}
        \xrightarrow{\pi(a|s)}
        \begin{array}{|c|c|c|}
            \hline
            . & . & . \\
            \hline
            . & x & . \\
            \hline
            . & . & . \\
            \hline
        \end{array}
        \xrightarrow{T(s'|s)}
        \begin{array}{|c|c|c|}
            \hline
            . & . & . \\
            \hline
            . & x & . \\
            \hline
            . & o & . \\
            \hline
        \end{array}
        \rightarrow \cdots
    \end{align*}
    \caption{An example fragment of the tic-tac-toe game, where the player $x$ goes first.}
    \label{fig:tictactoegame}    
\end{figure}

The game is played on a 3x3 grid board, each square being either empty or containing one of $x$, $o$. 
The model plays $x$, and opponent actions are sampled randomly across all moves. \Cref{fig:tictactoegame} depicts an example game progression.
The game ends when there are three of the same symbol in any row, column, or diagonal, or if there are no more possible moves. 
A reward of $1$ is given for winning the game, $0$ for losing the game, and $\frac{1}{2}$ for the draw. The player has the first move or goes second with equal probability. Formally, we can model the game as a Markov Decision Process. The transition function $\T(s'|a, s)$ first deterministically places the player's symbol in the chosen square and then samples the opponent's move.

\subsubsection{Experiment description}

We first train the model on a dataset consisting of all 255168 possible Tic-Tac-Toe games. We use an encoding similar to that introduced in \citet{chen_decision_2021}, which is given by a sequence of alternating states and rewards-to-go.  It can be thought of as a Decision Transformer trajectory, where we consider the actions to be board states after the \emph{player's} move, and the states to be the board state after the \emph{opponent's} move. We describe the exact textual encoding of the game in the~\Cref{appendix:tictactoe}.

We then condition the model on victory and measure its performance with an opponent playing uniformly at random. 
We record $10 000$ such trajectories, and then iteratively fine-tune the model on them, repeating the experiment four times and reporting the averages. Each result was computed on $2000$ games per value per experiment.
Note that the model always goes first in the fine-tuning training data.
Our theory predicts that the resulting model will get better at playing the game.
Note also that this does not imply that the model should become more coherent, except in the limit.

\begin{table}
    \centering
    \small
    \caption{Fine-tuning the decision transformer on a dataset consisting of its own games improves performance.}

    \renewcommand*{\arraystretch}{1.2}
    \begin{tabular}{lccccc}
        Iteration & P1 score & P2 score & P1 KL & P2 KL & \\
        \hline
        \hline
        Unconditioned & 0.5725 & 0.3325 & 0.0425 & 0.055\\
        1st Fine-tune & 0.725 & 0.3275 & 0.1025 & 0.03 \\
        2nd Fine-tune & 0.7675 & 0.3175 & 0.115 & 0.03\\
        \hline
        Conditioned & 0.7125 & 0.615 & 4.48 & 6.0625 \\
        1st Fine-tune & 0.8225 & 0.6775 & 4.0475 & 5.195 \\
        2nd Fine-tune & 0.8325 & 0.6825 & 4.625 & 5.6775
    \end{tabular}
    \label{tab:finetuning}
\end{table}

\subsubsection{Model and Results}
We train a predictive model using the \emph{gpt-mini} architecture of minGPT~\citep{andrej_karpathymingpt_2024}.
The model consists of six transformer decoder layers with six attention heads per layer and an embedding dimension of 192, for 2.68 million parameters in total. We report the results of the experiment in \Cref{tab:finetuning} as the average score where a win scores 1, a draw 0.5, and a loss 0 points. We see model performance improve with fine-tuning on a dataset containing games generated with the model. We also report the empirical KL divergence  $D_{KL}(\pi_{cond}||\pi)$ where $\pi_{cond}$ is the victory-conditioned predictive model, and $\pi$ is the unconditioned predictive model (as in~\Cref{def:coherence}).


\section{Discussion}
\label{sec:discussion}

Building on the work of \citet{ortega_shaking_2021}, we have described auto-suggestive delusions and introduced predictor-policy incoherence, two limitations of agents simulated by predictive models. Since even a perfect predictive model is still subject to both limitations, we argue that the problem cannot be avoided simply by increasing the model size or by supplying more observational data. However, we showed that online fine-tuning the model on its own outputs mitigates both problems, both formally, in the limit, and in practice, in our synthetic games datasets. 


\subsection{Limitations and future work}

Although we started to provide a unified framework for how why agents derived from predictive models can fail, there are still a lot of open questions. On the theory side, we have only formally considered single-step latent state setups, and only considered MDPs with finite time-horizon with finite state and action spaces. Our treatment of the mitigations of confounding is limited to the already-known solution of online fine-tuning. On the experimental side, one clear next step is to extend these results to the domain of full-scale language models. \citet{ortega_shaking_2021} suggests that auto-suggestive delusions are a component of the tendency of LM agents to hallucinate. Similarly, predictor-policy incoherence might offer a more formal explanation for why LMs require careful prompt engineering to achieve high performance, again supplementing the intuitive explanations already given by \citet{shanahan_role-play_2023}.

\subsection{Development of capabilities}

Our work has implications for both the type of task on which simulated agents might develop superhuman capabilities, and the kind of regime which might make this more likely. 

We should expect that even large models will fail to simulate agents that perform optimally on tasks where the model cannot be conditioned to specifically simulate competent agents, and tasks where the agents rely on information the model cannot access. In the case of predictor-policy incoherence we should expect models to simulate agents with more average capabilities, and in the case of auto-suggestive delusions, we should expect models to simulate agents which behave as if they have made nonexistent observations. In both cases, these limitations should fade as a model undergoes further training on its own outputs. One important next step is to investigate how methods like RLHF and DPO~\citep{christiano_deep_2017, bai_training_2022,rafailov_direct_2023} impact models' predictor-policy incoherence and agency, independently of the particular reward model.

\subsection{Scalable oversight through limiting agency}

Auto-suggestive delusions and predictor-policy incoherence limit the capabilities of agents simulated by predictors. Instead of fixing those limitations, we might want to embrace them as a tool for scalable oversight \citep{bowman_measuring_2022}. On almost all tasks, scaling a model leads to better performance. However, there are certain key capabilities we might want large models to lack because of their potential for harm or misuse. One approach could be to intentionally construct a training distribution which leverages predictor-policy incoherence and auto-suggestive delusions to have the model simulate powerful agents which are mistaken about their causal position and what features of the world they have causal control over, or what kinds of complex sequences of actions they would actually be able to successfully take, ultimately biasing the model towards simulating certain kinds of preferred policies.


\bibliography{bibliography}
\bibliographystyle{icml2024}

\newpage
\appendix
\onecolumn

\section{Policy incoherence proof}\label{appendix:policy-confounding-proof}

\begin{theorem}
    If a policy $\pi_0$ has full support over the set of actions $A$ in all states $S$, then iterating the operator $\mathcal{G}$ gives an optimal policy. That is, there exists some optimal policy $\pi^*(a|s)$, such that the sequence of policies $\pi_k(a_t|s_t)$ converges to $\pi^*$ in $KL$-distance:
    \[
    \lim_{k \to \infty} D_{KL}(\pi_k||\pi^*) = 0
    \]
    where $\pi_k = \mathcal{G}(\pi_{k-1}) = \mathcal{G}^{k}(\pi_0)$.
\end{theorem}

\begin{proof}
    The proof is by induction on $T$. Our induction hypothesis is going to be stronger: we will show an explicit characterisation of the limiting policy $\pi^*(a_t|s_t)$. Let us first define the set of action-trajectories $A^T = \{(a_1, a_2, \ldots a_T)\}$, and the subset of optimal action trajectories:
    \[
    A^{T*} = \left\{a_{1:T} \in A^T : p(R | a_{1:T}) = \max_{a'_{1:T} \in A^T} p(R|a'_{1:T})\right\}
    \]
    where $p(R|a_{1:T})$ is the appropriately marginalized probability distribution $p$. After projecting on the $t$-th dimension we get sets $A^{t*} = \{a_t : (a_1, \ldots a_T) \in A^{T*}\}$. Now we claim that the limiting policy $\pi^*(a_t|s_t)$ is going to be:
    \[
        \pi^*(a_t|s_t) \ \propto \ \mathbbm{1}_{[a_t \in A^{t*}]} \pi_0(a_t|s_t)
    \]
    
    Let us first assume that $T = 1$. Then, since the starting state is assumed to be deterministically chosen $s = s_1$, the distribution $p$ simplifies to (from definition):
    \[
    p(s, a, R) = \pi(a|s)p(R|a)
    \]
    where the notation $p(R|a) = p(R = 1|a, s)$ is unambiguous since the reward is binary and the state $s$ is known. Therefore, we can prove by induction on $k$ that:
    \[
    \pi_k(a_t|s_t)\ \propto\ \pi_0(a_1|s_1)p(R|a_1)^{k}
    \]
    Base case $k = 0$ is trivial. Now for $k > 0$, expanding the definition and applying Bayes law we get:
    \begin{align}\label{eq:bayes-law-iteration}
        \pi_k(a|s) &= p_k(a|s, R = 1) 
        \\
        &= \frac{p_{k}(R = 1|s, a) p_{k}(a|s)}{ p_{k}(R = 1|s)} \notag \\
        &\ \propto\ \pi_{k-1}(a|s)p(R|a) \notag \\
        &=  \pi_0(a|s)p(R|a)^{k}\notag
    \end{align}
    where the last line follows from the induction hypothesis.  Therefore, for any two actions $a_1, a_2$, we have that:
    \[
    \frac{\pi_k(a_1|s)}{\pi_k(a_2|s)} = \frac{\pi_0(a_1|s)}{\pi_0(a_2|s)}\cdot\left(\frac{p(R|a_1)}{p(R|a_2)}\right)^k
    \]    
    Let us take the set of optimal actions:
    \[
    A^* = \left\{a \in A: p(R|a) = \max_{a^* \in A}p(R|a^*)\right\}
    \]
    We then know that for any action $a \notin A^*$ we have:
    \begin{equation}\label{eq:limit-iteration}
            \lim_{k \to \infty}\pi_k(a|s) \leq \lim_{k \to \infty}\frac{\pi_k(a|s)}{\pi_k(a^*|s) + \pi_k(a|s)} = \frac{1}{\frac{\pi_k(a|s)}{\pi_k(a|s)} + 1} = 0
    \end{equation}
    
    where $a^*$ is some action in $A^*$. This implies the optimality: the fact of the policy being optimal is equivalent to having support contained in $A^*$. We can apply \Cref{eq:limit-iteration} to a $a', a'' \in A^*$ - their likelihood ratio is then simply constant:
    \[
    \lim_{k \to \infty}\frac{\pi_k(a'|s)}{\pi_k(a''|s} = \frac{\pi_0(a'|s)}{\pi_0(a''|s)}
    \]
    since they both give the same chance of reward. Thus, the process of iterating $\mathcal{G}$ converges simply to the desired distribution $\pi^*(a|s) \ \propto \ \mathbbm{1}_{[a \in A^*]} \pi_0(a|s)$.
    
    Now, we move to the induction step and assume $T > 1$. Let us consider the first action of the policy $\pi_0(a_1|s_1)$ at $t = 1$. Because of the Markov property and the fact that reward is only given at $t = T$, that is, $p(R|s_1, a_1, s_t) = p(R|s_t)$, we have the equation:
    \[
        p(a_t|s_1, a_1, s_t, R=1) = p(a_t|s_t, R=1)
    \]
    Therefore, the value of $\pi_k(a_t|s_t)$ for any $t > 1$ is computed independently of $\pi_k(a_1|s_1)$. This allows us to use the inductive hypothesis on the Markov chain restricted to $t \geq 2$, of length $T-1$. This establishes the right limits for all $\pi^*(a_t|s_t)$ for $t \geq 2$, leaving only the question of $\pi^*(a_1|s_1)$. We then note that:
    \[
    p_k(R|a_1) = \mathbb{E}_{s_2 \sim \tau(s|s_1)}[p_k(R|s_2)]
    \]
    where the RHS is again independent of any of $\pi_i(a_1|s_1)$, and converges simply to the fixed value:
    \[
    \lim_{k \to \infty} p_k(R|a_1) = \max_{a_{2:T} \in A^T} p(R|s_2, a_{2:T}) 
    \]
    Now, we use the same argument as in~\Cref{eq:limit-iteration} to show that for any $a_1 \notin A^{1*}$ we have:
    \[
    \lim_{k \to \infty} \pi_k(a_1|s_1) = 0
    \]
    and similarly for any two $a_1', a_1'' \in A^{1*}$ we have the ratio:
    \[
    \lim_{k \to \infty} \frac{\pi_k(a'_1|s_1)}{\pi_k(a''_1|s_1)} =  \frac{\pi_0(a'_1|s_1)}{\pi_0(a''_1|s_1)}
    \]
    which completes the proof. Since $\pi_k(a_t|s_t)$ converge in distribution to $\pi^*(a_t|s_t)$ (point-wise in each $s_t$), they also converge in KL distance.

\end{proof}

\newpage
\section{Tic-tac-toe experiment details}
\label{appendix:tictactoe}

\subsection{Textual encoding}
To encode the games, we use a vocabulary of 8 tokens of $x, o, ., \#, O, D, \$, E$, where the first three indicate positions on the board, the next three indicate who won (crosses, noughts, draw) and the last two are BOS and EOS tokens respectively. Each game trajectory is a sequence of tokens of a form depicted in \Cref{fig:tictactoeencoding}. We use context window of 30 tokens, which is exactly three states. This means that there are always at least two full board positions visible in the context window, which makes it possible to determine whose turn it is. BOS tokens are padded so that the initial context window contains only tokens up to the end of the first (empty) state. We pad EOS tokens so that each game's length is equal.

\begin{figure}[ht]
    \centering
    \[
    \underbrace{\$\ldots\$}_{\substack{\text{padded} \\ \text{BOS}}}
    \underbrace{\#}_{\substack{\text{game} \\ \text{outcome}}}
    \underbrace{.........}_{\substack{\text{game} \\ \text{start}}}
    \underbrace{\#}_{\substack{\text{game} \\ \text{outcome}}}
    \underbrace{....x....}_{\substack{\text{player's} \\ \text{move}}}
    \underbrace{\#}_{\substack{\text{game} \\ \text{outcome}}}
    \underbrace{....x..o.}_{\substack{\text{opponent's} \\ \text{move}}}
    \ \ \ 
    \ldots
    \]
    \caption{Textual encoding of the part of the game depicted in \Cref{fig:tictactoegame}, assuming that the cross player eventually wins.}
    \label{fig:tictactoeencoding}
\end{figure}

\begin{figure}[h]
    \centering
    \includegraphics[width=6cm]{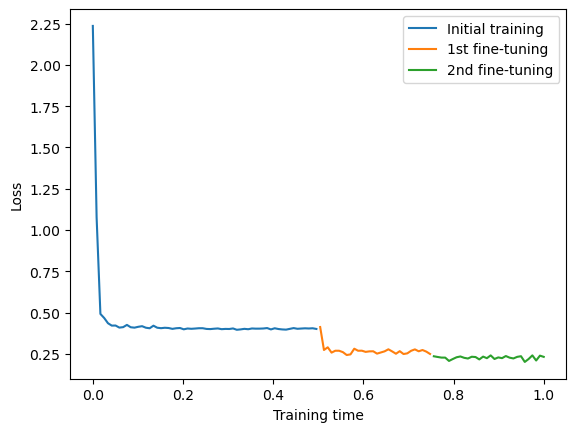}
    \caption{Training loss in the tic-tac-toe game. Subsequent fine-tuning epochs correspond to the visible drops in the loss.}
    \label{fig:tictactoeloss}
\end{figure}

\begin{figure}%
    \centering
\subfloat{{\includegraphics[width=6.5cm]{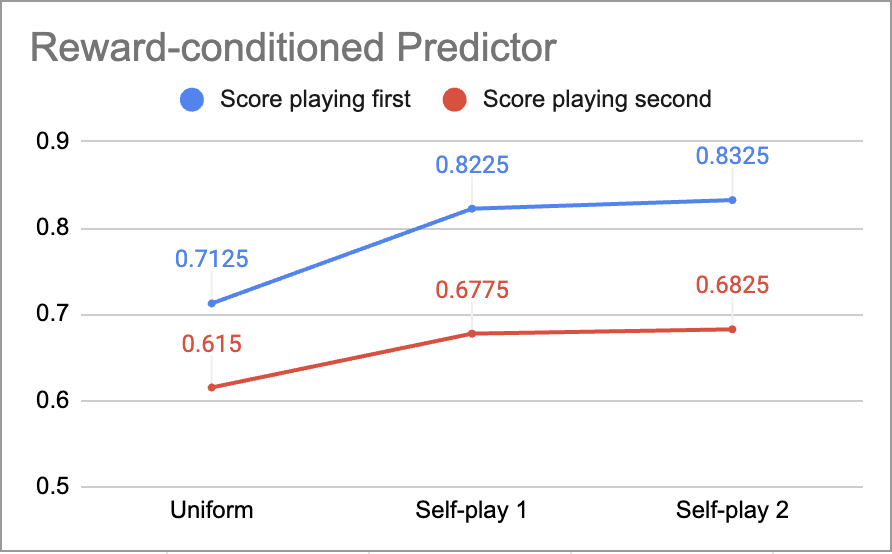} }}%
    \qquad
\subfloat{{\includegraphics[width=6.5cm]{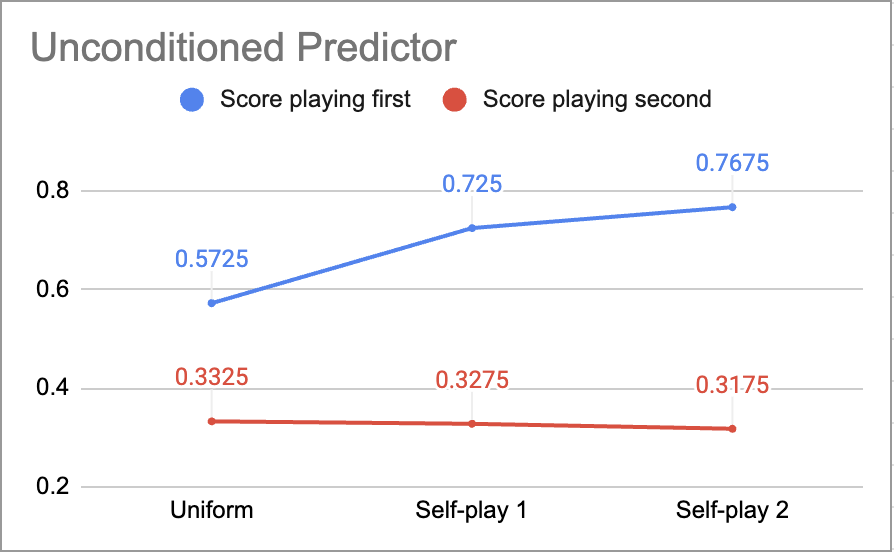} }}%
    \caption{Scores of iteratively fine-tuned conditioned and unconditioned predictors in the tic-tac-toe experiment.}%
    \label{fig:example}%
\end{figure}

\newpage
\section{Padlock experiment details}
\label{appendix:padlock}

\begin{figure}
\centering
\includegraphics[width=0.47\textwidth]{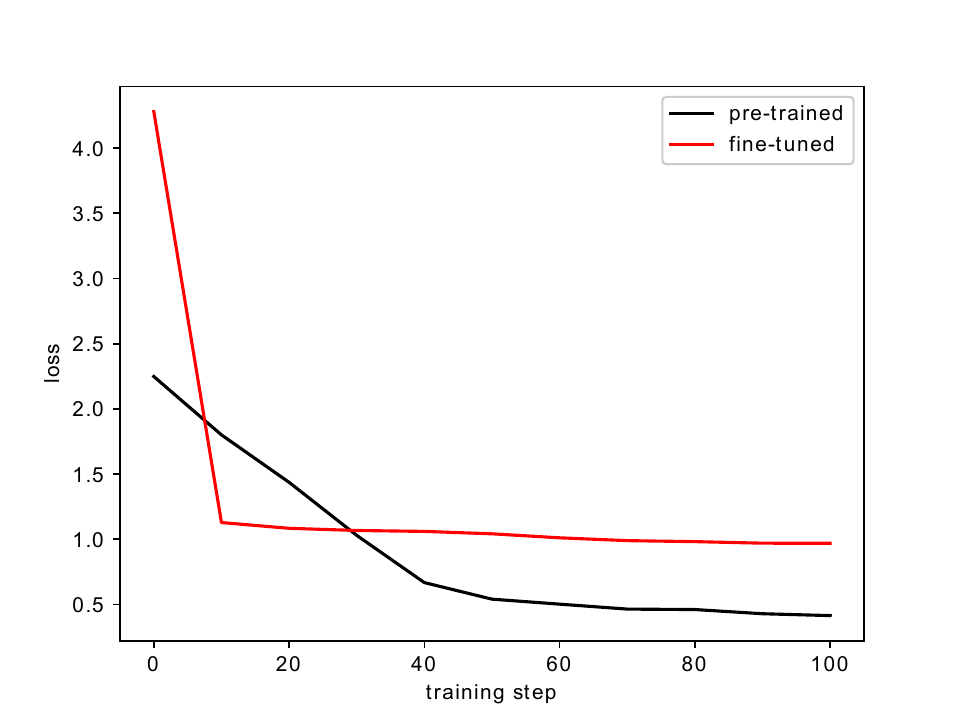}
\caption{Test loss depending on the training step.}
\label{fig:padlock_training}
\end{figure}

See Figure \ref{fig:padlock_training} for the training dynamics on a simple version of the padlock task with three fixed and three unknown bits for the target configuration. It compares pre-training on the dataset of expert play to fine-tuning the pre-trained model on the outputs of the pre-trained model. The fine-tuning loss converges to a higher value than the pre-training loss. This is as expected, because in the fine-tuning dataset there should be a larger uncertainty whether an action flipping one of the last three levers will open the lock. This uncertainty is higher because in the fine-tuning dataset the model might need to go through all 8 combinations of the last three levers, whereas in the pre-training dataset experts know which ones to flip, resulting in only four combinations: from no levers flipped to all three being flipped.

This discrepancy between the uncertainties over the lever combinations is reflected by the delusion measure. We implemented the delusion measure to estimate the degree to which the model exhibits auto-suggestive delusions. Specifically, after the model has taken actions for the first three levers, we sample its probabilities for the success token "S" in the remaining positions. As a delusion measure, we calculate the KL divergence of these probabilities with the ideal probability distribution for this task.

For example, after the first three levers are pulled, a pre-trained model would predict with close to 100 percent probability that the next token is "S". That can not be correct as only 4 out of 8 combinations have been tried. If the initial state of the last three levers is 000, then these combinations would have been tried by sequentially pulling the last three levers: 000, 100, 110, 111.

In this case, having nearly 100 percent overconfidence in the environmental state of the lock opening is an example of an auto-suggestive delusion. The pre-trained model observes its past actions of pulling the last three levers sequentially and effectively updates its belief that a latent state $\theta$ is such that all three of the last levers must be pulled to unlock the lock -- this would be a valid inference when predicting experts. This leads the model to an auto-suggestive delusion of the next token being the success token.

Generally, the model that is fine-tuned is more likely to open the lock. This is because the fine-tuned model has observed when the pre-trained model failed or succeeded when interacting with the padlock. When conditioned on opening the padlock, this allows the fine-tuned model to select better strategies such as trying out more lever combinations. Conditioning on opening the lock is technically implemented by adding a token indicating success or failure to the beginning of the sequence and then setting it to success before predicting actions.

\newpage

\section{Fixing observational confoundedness by fine-tuning agents on their own outputs}\label{appendix:delusion-proof}

Here we give the proof that a predictive model will stop being confounded by hidden observations after re-training on feedback from its own actions. It is considerably simpler than the case of policy confounding and incoherence.

We'll start by considering a very simple game in which an expert makes a single observation, takes an action, and then reaches an end state.

Suppose that there is some initial distribution $P(o)$ over observations, and true transition function $\tau(s | a, o)$, as well as some expert policy $\pi(a | o)$ that generated the data.

When we train the predictive model $Q$ on this data, without the observations, all it will learn is some $Q(a)$ and $Q(s | a)$, which will indeed be the true probabilities of an expert taking an action, and of a state following from that agent’s action. $Q(s | a)$ is effectively the true probability of transition $\tau(s | a, o)$ marginalised over $o$ given the policy $\pi(a | o)$: $$\sum_{o} \tau(s | a, o) P(o) \frac{\pi(a | o)} {P(a)}$$

But if we use the predictive model $Q$ to simulate agents and predict the consequences of their actions, the actions will be independent of the observations. This is where the confounding arises: the simulated agent’s policy $\pi^*(a)$ will just be $P(a)$. So the \textit{actual} probability of a state resulting from the simulated agent’s actions will be 
$$\sum_{o} \tau(s | a, o) P(o)$$ 
However, data generated using $Q$’s simulated agent and real feedback exhibits confounding precisely because $a$ and $o$ are independent. So a model re-trained on this data will correctly learn 
$$Q(s | a) = \sum_o \tau(s | a, o) P(o)$$

This result generalises to scenarios with multiple actions, observations, and states. For a given action $a$, if we take $h$ to be the history of the game observable to the predictive model, and $\mathbf{o}$ to represent any hidden observations available to the expert which are not conditionally independent of $s$ given $h$ and $a$, we again want the model to learn 

$$Q(s | a, h) = \sum_\mathbf{o} \tau(s | a, \mathbf{o}, h) P(\mathbf{o} | h)$$

Which is indeed the distribution in data generated by its own simulated agents.

Note that to generate this data, we do not require direct access to the hidden observation, but we do need to be able to sample what state results from a given action, assuming an independently generated observation, so we need to be able to guarantee that the hidden observation has been randomised, take an action, and observe the resultant state.


\end{document}